\documentclass[11pt]{amsart}
\pdfoutput=1
\usepackage{amsmath,amssymb,amsthm}
\usepackage{enumitem}
\usepackage{tikz}
\usepackage{tikz-cd}
\usetikzlibrary{decorations.pathmorphing}
\usepackage{bbm}
\usepackage{array}
\usepackage{hyperref}
\hypersetup{
    colorlinks=true,
    linkcolor=blue,
    filecolor=magenta,      
    urlcolor=blue}
\usepackage{xcolor}
\usepackage{graphicx}

\theoremstyle{plain}

\numberwithin{theorem}{section}

\theoremstyle{definition}

\newtheorem*{rep@theorem}{\rep@title}
\newcommand{\newreptheorem}[2]{%
\newenvironment{rep#1}[1]{%
 \def\rep@title{#2 \ref{##1}}%
 \begin{rep@theorem}}%
 {\end{rep@theorem}}}

\newreptheorem{theorem}{Theorem}
\newreptheorem{lemma}{Lemma}
\newreptheorem{proposition}{Proposition}
\newreptheorem{definition}{Definition}
\newreptheorem{corollary}{Corollary}

\title{Toward Errorless Training ImageNet-1k}

\author[Deng]{Bo Deng}
\address{
  \begin{tabular}{l}
   Bo Deng \\
   \hspace{.1in} University of Nebraska-Lincoln \\
      \hspace{.1in} Department of Mathematics \\
   \hspace{.1in} 318 Avery Hall, 1144 T St, Lincoln, NE 68588-0130\\
   \hspace{.1in} Email: {\bf bdeng1@nebraska.edu} \\
  \end{tabular}
}

\author[Heath]{Levi Heath}
\address{
  \begin{tabular}{l}
   Levi Heath \\
   \hspace{.1in} University of Colorado Colorado Springs \\
      \hspace{.1in} Department of Mathematics \\
   \hspace{.1in} 1420 Austin Bluffs Pkwy, Colorado Springs, CO 80918\\
   \hspace{.1in} Email: {\bf lheath2@uccs.edu} \\
  \end{tabular}
}

\begin{document}

\maketitle

\noindent
\textbf{Abstract}: In this paper, we describe a feedforward artificial neural network trained on the ImageNet 2012 contest dataset \cite{ImageNet1kData} with the new method of \cite{deng2023error-free} to an accuracy rate of 98.3\% with a 99.69 Top-1 rate, and an average of 285.9 labels that are perfectly classified over the 10 batch partitions of the dataset. The best performing model uses 322,430,160 parameters, with 4 decimal places precision. We conjecture that the reason our model does not achieve a 100\% accuracy rate is due to a double-labeling problem, by which there are duplicate images in the dataset with different labels.  


\section{Introduction}

Artificial intelligence (AI), and intelligence in general, should exhibit these traits at a minimum: predictability, creativity, and learning.  For AI models built on neural networks, the predictability is based on the continuity of all transformations in the networks as building blocks. Namely, every parameter and variable has its neighborhood of continuity in which the model behaves similarly, leading to the theoretical basis for predictability. All large language models are capable of hallucinating, and hence are creative by definition. The third capability, however, can be defined rigorously. By definition, learning is not making the same mistake twice. In practice, this means that a model can be trained error-free every time. This criterion is both necessary and sufficient. In fact, for discrete functions, errorless training has been guaranteed theoretically by the discrete version of the Universal Approximation Theorem (UAT, Theorem 2.5 of \cite{Hornik1989}) since 1989. 

However, no methods for errorless training were known til 2023, when the first author presented a new algorithm, called \emph{gradient descent tunneling} (GDT) in \cite{deng2023error-free}. The method uses any stochastic gradient descent (SGD) algorithm to train a feedforward neural network (FNN, \cite{rosenblatt1958perceptron,ackley1985learning}) to a modest accuracy and then follows it up by training the model to zero error rate one data batch at a time, using the idea of homotopy (\cite{chow2006homotopy}) from Numerical Analysis and Dynamical Systems, fields far from Machine Learning (ML). To demonstrate the effectiveness of GDT, \cite{deng2023error-free} trained several error-free models on the MNIST benchmark data for classifying handwritten digits \cite{lecun1998convolutional, lecun2002gradient}. The model weights for error-free FNNs are provided in the GitHub repository \cite{GitHubRepo_Error-Free_AI_Models} for the MNIST, as well as for the FashionMNIST (\cite{lecun2002gradient, FashionMNIST}) by the second author. These datasets are relatively small in size. The question is, can GDT be effective for large data sets? An affirmative answer to this question has practical import for predictive precision AI.  

The purpose of this paper is to report the results of applying the GDT method to the ImageNet-1k dataset \cite{ImageNet1kData}. We will present a simple model architecture with the aim of training the dataset error-free or, at the very least, achieving high accuracy rates. The model architecture is similar to a mixture of experts model \cite{shazeer2017outrageously}, consisting of multiple FNN feature modules in parallel. We will present evaluation results on the model's performance, which is independent of training, and evaluations on the model's training, which only requires the model's weight and bias parameters. The highest performing model has 98.3\% accuracy with a 99.969\% Top-1 accuracy rate, and among its constituent modules, the lowest accuracy rate is 98.965\% and the highest accuracy rate is 100\%. There are 285.9 labels on average for each of the 10 batches of the dataset that achieve the 100\% accuracy rate. The best performing model uses 322,430,160 parameters and 1,606,160 neurons. Our results also suggest that the reason for our model's imperfect performance is due to possible labeling inconsistencies in the ImageNet-1k dataset. Specifically, images may appear multiple times in the dataset with different labels.

Before our model, according to the benchmark tracking website Papers with Code, \cite{yu2022coca} developed the most accurate model trained on the ImageNet-1k dataset with $91.0\%$ accuracy and 2.1 billion parameters. The model with the fewest parameters was \cite{fontana2024distilled}, which attained $65.59\%$ accuracy and 1.03 million parameters. See also \cite{krizhevsky2012imagenet} for the AlexNet, which achieved a Top-1 37.5\% error rate with 60 million parameters and 650,000 neurons. 

With the utilization of GPUs, many advancements in model architecture have been made that greatly increase the accuracy of machine learning models, e.g.,  \cite{recht2011hogwild,  krizhevsky2014OneWeirdTrick}, and many others. Still, the approach has largely remained the same: do some versions of stochastic gradient descent until the model reaches a sufficiently accurate benchmark (\cite{rumelhart1986learning,lecun1998convolutional,lecun2002gradient,karpathy2014large}). We believe that our approach presented in this report should provide us with a novel training alternative in the fields of ML and AI. 

\section{Model Architecture}

A model of image classification, $M$, is a mapping that takes in an image as input and outputs a label. 
Our model is structured as follows. First, a fixed number of features of the image are selected for a model. For example, for a color image, a feature can be one of the three primary color channels, red (R), green (G), and blue (B), or a combination of the RGB channels, such as the Y channel $Y=0.2126R+0.7152G+0.0722B$. A combination of selected $n$ features uniquely defines a model, referred to as a featured model.  Seventeen different features are listed in Tab.\ref{tab_feature}. The set of the RG weights for RGg1,  0.618 and 0.382, comes from the Golden ratio. This choice was based on observing that the primary color features are generally easier to train than the features with equal weights -- e.g., RG and eRGB. RGg2 is the symmetric feature of RGg1. Feature BW is a popular choice for the luminance of color images. The RGB color space conversion to the XYZ color space is given by the last $3\times 3$ matrix.

\begin{table}[t]
    \centering
    \begin{tabular}{l|rrr}
    feature & \multicolumn{3}{|c}{RGB Weights}\\
    \hline
     R    & 1       &  0       &        0 \\
     G    & 0       &       1  &        0 \\
     B    & 0       &       0  &        1 \\
     RGg1 & 0.618   &   0.382  &        0 \\
     RBg1 & 0.618   &       0  &    0.382 \\
     GBg1 & 0       &   0.618  &    0.382 \\
     RGg2 & 0.382   &   0.618  &        0 \\
     RBg2 & 0.382   &   0      &    0.618 \\
     GBg2 & 0       &   0.382  &    0.618 \\
     RG   & 0.5     &     0.5  &        0 \\
     RB   & 0.5     &     0    &      0.5 \\
     GB   & 0       &     0.5  &      0.5 \\
     eRGB & 1/3     &     1/3  &      1/3 \\ 
     BW   & 0.299   &   0.587  &    0.114 \\
     X    & 0.4125  &  0.3576  &   0.1804 \\
     Y    & 0.2126  &  0.7152  &   0.0722 \\
     Z    & 0.0193  &  0.1192  &   0.9502 \\
    \end{tabular}
    \vskip .2in
    \caption{RGB Features}
    \label{tab_feature}
\end{table}

Next, we subject every feature of the model to a determination of labeling. To do so, we partition the set of labels into $k$ groups, which are referred to as modules. For the ImageNek-1k dataset, we partition the 1000 labels into $k=40$ (primary) modules, each containing 25 labels. For example, module one ($m_1$) consists of labels from 0 to 24, and $m_{40}$ consists of labels from 960 to 999.  

\begin{figure}[t]
\centerline
{\scalebox{.4}{\includegraphics{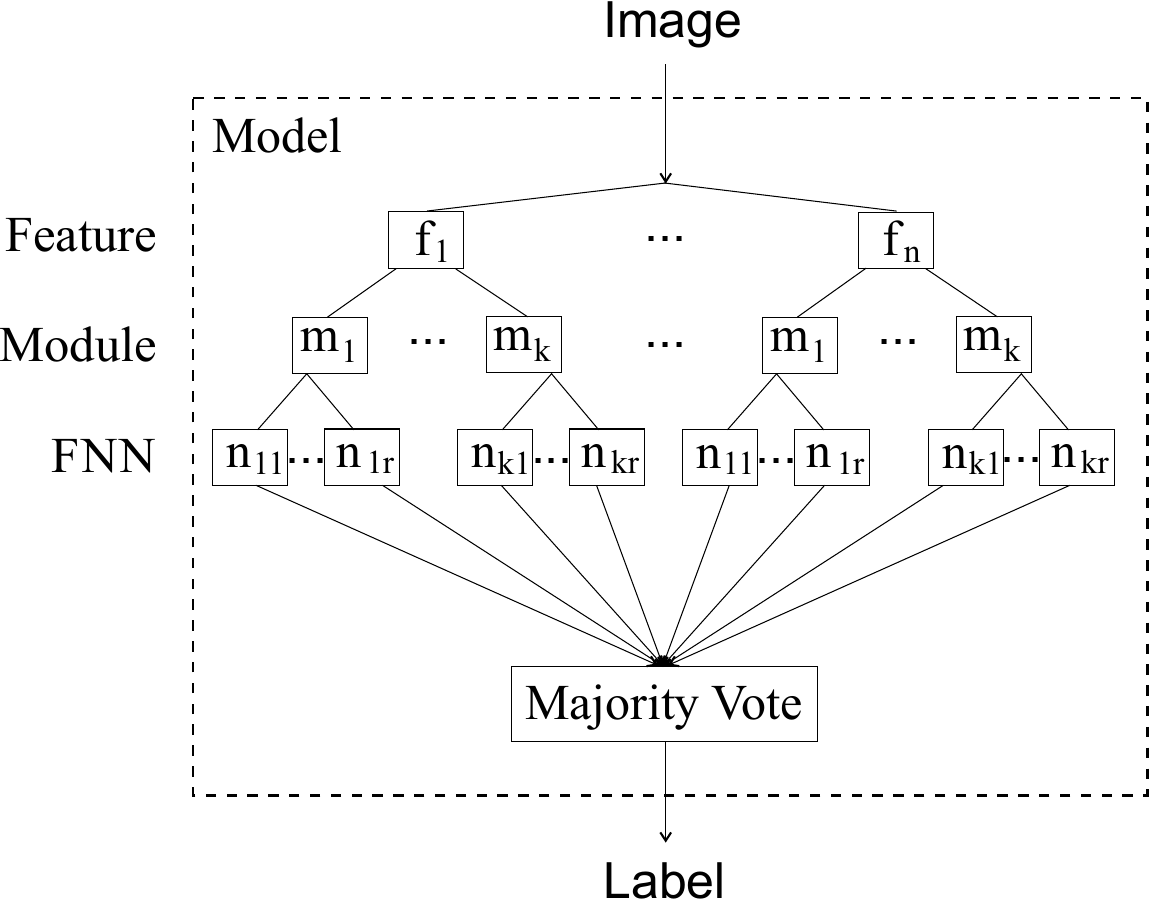}}}
\caption{\text{Model Architecture.} Every branch from the input to the output is an FNN. Each branch is uniquely defined by the triplet $(i,j,k)$ for feature $f_i$, module $m_j$, and sub-module $n_{js}$. The corresponding FNN is $N_{i,j,s}$ with transformed training batch $B_{i,j,s}$. The model is a parallel collection of $n\times k\times r$ FNNs.}
 \label{fig_model}
\end{figure}

For each module $m_j$, we train $r$ FNNs, $n_{j1},\dots, n_{jr}$, see Fig.\ref{fig_model} for an illustration of the architecture. For the first module $m_1$ of labels numerated from 0 to 24, the output of $m_1$'s $r$ FNNs, $n_{1s}$, is a label between 0 and 24. We will use the cross-entropy for the loss function and the nearest-neighbor protocol for classification.  

As a result, for every input image, each FNN will give a prediction in its module labels with the minimum cross-entropy loss. In total, there are $n\times k\times r$ predicted labels, together with their nearest-neighbor cross-entropy losses. We will denote the labels and the losses by $[\ell_{i,j,s},\ \varepsilon_{i,j,s}]$ with $1\le i\le n$ for the feature number, $1\le j\le k$ for the module number, and $1\le s\le r$ for the FNN number. 

The last operation of the model applies the majority voting method to determine a final prediction from the collection of predicted labels $\{\ell_{i,j,s}\}$ by finding the minimum of the losses $\{\varepsilon_{i,j,s}\}$. In the case of a tie, we use the smallest variance of the losses to break the tie. Because the values of the losses are real numbers with many decimal places, the smallest variance is almost surely unique.

To complete the description of the model architecture, we use a standard FNN with one or two hidden layers for all $n_{js}$ of all features. The number of nodes for the output layer is $1000/k$, which equals 25 in our case since we take $k=40$. The first hidden layer has 256 nodes, and the second, if present, has 77 nodes. The number of nodes for the input layer is 900, which we explain in \S\ref{sec-training}. As a result, we use $900\times 256\times 25$ architecture for FNNs with one hidden layer, and $900\times 256\times 77\times 25$ for FNNs with two hidden layers. 

Notation-wise, we will use S\_h$x$\_m$y$ for SGD trained models and T\_h$x$\_m$y$ for GDT trained models, with all 17 RGB features described in Tab.\ref{tab_feature}. To distinguish the usage of featured models later, we will call them \emph{proto-model}. Here, $x$ with $x=1,2$ is for the number of hidden layer(s),  and $y$ with $y=1,2$ is for different proto-models. That is, the first capital letters, S or T, are for the type of training, and the integer index $y$ is for distinct proto-models trained by the methodology S or T, with the hidden architecture h$x$. For example, S\_h2\_m1 is the first SGD proto-model with 2 hidden layers. With one hidden layer, each proto-model with $n=17,k=40,r=2$ has $17\times 40\times 2\times(900+256+25)=$ 1,606,160 neurons and $17\times 40\times 2\times(901\times 256+257\times 25)=$ 322,430,160 many parameters for edge weight matrices $W$ of sizes $900\times 256$ and $256\times 25$ and bias vectors $b$ of sizes $256\times 1$ and $25\times 1$ for the hidden layer and the output layer respectively.   

\section{Training}\label{sec-training}

Training is carried out for the proto-models. There are more than 1.2 million images in the ImageNet-1k dataset, with 1000 labels. To train all FNNs, we first organize the images into $k$ batches, with $k$ the same as the number of modules. Images from a batch have the same labels as those that are used to train the corresponding module. For example, batch 1 contains all images with labels from 0 to 24, and batch 40 contains all images with labels from 960 to 999. On average, each batch contains about 30,000 images. The reason for having multiple FNNs for each module/batch is for easier training of the FNNs. In our case, we use $r=2$ FNNs for each module/batch so that FNN $n_{j1}$ is trained on the first half images of the batch $m_j$, and FNN $n_{j2}$ is trained on the second half images of the batch, each containing about 15,000 images. Therefore, every feature of every labeled image from the ImageNet-1k training set is trained at a unique FNN node of the model diagram Fig.\ref{fig_model}. Thus, we can use notation $B_{j,s}$ to denote the batch, corresponding to label module $j$, $1\le j\le k$, and subset batch $s$, $1\le s\le r$. The partition of the training set applies to all features, $f_i$, $1\le i\le n$. As a result, in terms of feature $f_i$, the input image batch $B_{j,s}$ is transformed into the corresponding feature, which we denote as $B_{i,j,s}$. We can also denote the corresponding FNN by $N_{i,j,s}$, which is trained on the featured image batch $B_{i, j,s}$. That is, each FNN $N$ is independently trained on feature $f$ of its training batch $B$.    

For all FNNs, $N_{i,j,s}$, the ReLU activation function is used for the hidden layer, the softmax is used for the output layer, and the cross-entropy loss function is used for the classification of FNN's output. As for the FNN parameters in edge weights, we will use 
$W^\ell_{i,j,s}$ for $N_{i,j,s}$ with superscript $\ell=1,2$ for the first and second layers. The same notation convention, $b^\ell_{i,j,s}$, applies to the bias parameters of the FNN models. $L_{i,j,s}$ denotes the cross-entropy loss function of the FNN $N_{i,j,s}$. For a deeper FNN, $\ell=1,\dots,q$, for $q>2$.  

By definition, to train the FNN model, $N_{i,j,s}$, is to find the global minimum of the loss function, $L_{i,j,s}$, at which every training image from $B_{i,j,s}$ is perfectly trained with the ground-truth label equal to FNN's predicted label. We achieve this goal in two steps. In step one, we use a stochastic gradient descent (SGD) method to find a local minimum of the loss function. In step two, we use the gradient descent tunneling method (\cite{deng2023error-free}) (GDT) to move from the SGD local minimum in the $\{W,b\}$ parameter space to a global minimum. In the end, we obtain two types of trained models: the SGD-trained proto-model, S\_h$x$\_m$y$, represented by its local minimum $\{W_s,b_s\}$, and the GDT-trained proto-model, T\_h$x$\_m$y$,  represented by its global minimum $\{W_t,b_t\}$. 

If the images, $B_{i,j,s}$, are self-consistent, namely, the labeling of images defines a function, the GDT method guarantees to find the global minimum, at which every image is correctly predicted by the FNN, $N_{i,j,s}$. If, on the other hand, the image labeling of the batch $B_{i,j,s}$ fails to be a function, then there exists at least one image that appears in $B_{i,j,s}$ at least two times with two different labels, i.e., a problem of double labeling. When this inconsistency occurs, the FNN is impossible to reach 100\% accuracy. As a result, the GDT method will find a parameter point in $\{W,b\}$ at which the accuracy is below 100\%. In other words, whenever the GDT-trained error rate is not zero, double labeling is suspected. 

\section{Evaluation}

Once the proto-models are trained, their FNN parameters $\{W,b\}$ completely define the models. Two types of evaluation can be done: one is referred to as the training evaluation, and the other is referred to as the model evaluation. For the first type, it is straightforward: evaluate the performance of each FNN in terms of its accuracy rate on its assigned training data. In particular, whether the GDT-type rate achieves 100\% accuracy or the training data contains contradictions. For the second type, all that matters is the proto-model defined by the parameters, regardless of how the parameters are obtained, for which the model's architecture becomes important. Evaluation compares the predicted label by the proto-model of all FNN modules with the true label for every training image. In summary, training evaluation compares the true label only with the predicted label by the designated FNN module for the image, whereas model evaluation compares the true label with the predicted label from all the FNN modules by the majority-voting protocol.    

In the ideal situation, if the entire training set of images is free of double labeling, then every image is perfectly trained for every feature. Thus, at the proto-model's output node for model evaluation, there will be at least $n$ correct predictions, all having the same ground-truth label. This takes place at the same module number $j$ and the same subset number $s$ for all features. For any combination other than the intended FNN, $N_{i,j,s}$, for all $1\le i\le n$, the prediction can be any label from 0 to 999. By chance, these predictions can achieve a majority equal to the possible maximum, the number of features, $n$, referred to as the super-majority number. However, the larger the number of features $n$ is, the less likely a chancy super-majority becomes. 

In fact, it is straightforward to see that if each feature is trained on only one module with $k=r=1$, then perfectly trained FNNs, $N_i:=N_{i,1,1}$, automatically lead to a perfectly trained model for all $1\le i\le n$, because majority voting will always output the true label, $n$ times, with the super-majority. If $k> 1$ or $r> 1$, then the aggregated model may not reach the 100\% accuracy rate, because every training image is also subject to FNNs not trained on the image, which can result in a false label. In fact, when $r=1$ and $k>1$, the FNNs of modules, to which the image does not belong, always result in a false label, which can form a super-majority by chance.  Therefore, the more features a model has, the higher its majority-ruled accuracy rate becomes, because it is less likely that false labels form a voting majority.     

\begin{figure}[t]
\centerline
{\scalebox{.5}
{\includegraphics{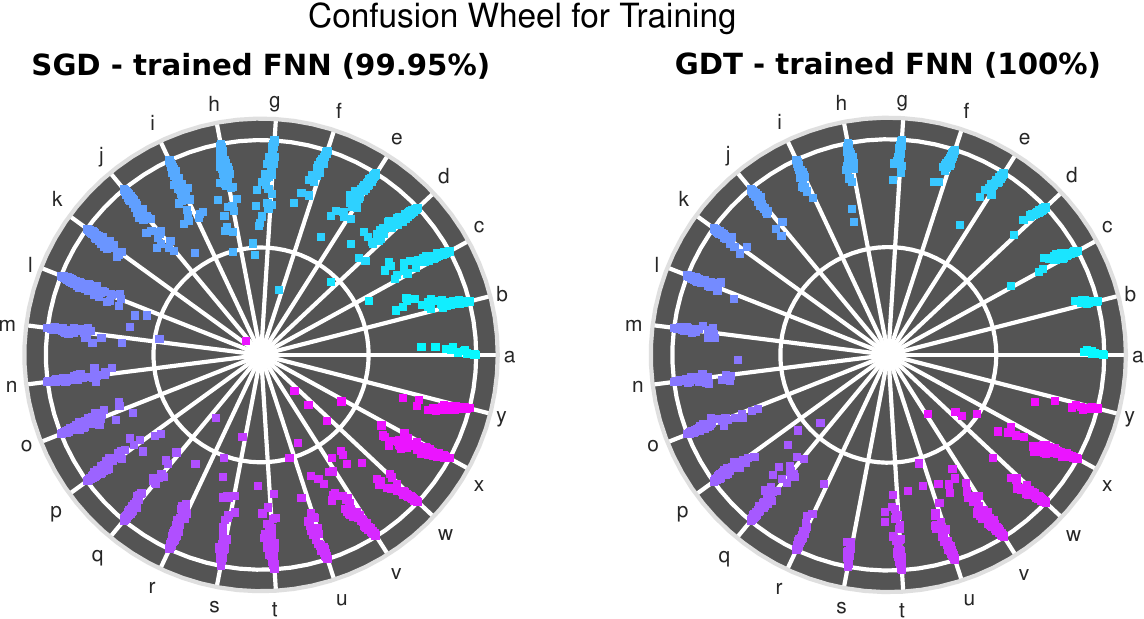}}}
\caption{\text{Confusion Wheel.} The plot is for the proto-model 
\text{S\_h2\_m1 and T\_h2\_m1}, with module $m_9$, sub-module $n_{92}$, and feature $f_4={\rm RGg1}$, namely, FNN $N_{4,9,2}$. The 25 labels of the featured-module FNNs are represented alphabetically. Labeled output vectors of the FNNs are linearly transformed to corresponding spokes in the unit circle. A misclassified point lies outside its designated sector with the corresponding centering spoke. The linear transformation matrix consists of the spoke vectors on the unit circle for its column vectors.}
 \label{fig_wheel}
\end{figure}

To illustrate this analysis, we will use six featured models in the section below, each having a different feature combination, all derived from the proto-models. For Model-1, we use the first three features of Tab.\ref{tab_feature}, i.e., the primary color channels of the images. Model-2 consists of the first six features of Tab.\ref{tab_feature}, and the next of the first nine, and etc, up to Model-6, having all seventeen features from the table.   

\section{Results}
\noindent
\textbf{Training Evaluation.} In the Hugging Face repository \cite{Repo_Towards_Errorless_Training_ImageNet-1k}, in the folder of Data\_Transformationfig, the code, Run\_Modularize\_Data.m reorganizes the dataset into $k=40$ modules, and then partitions each module into $r=2$ groups of training images. For the training accuracy rate, the code, Run\_Training\_Evaluation\_Model.m, evaluates the accuracy for the SGD-type and the GDT-type FNNs, $N_{i,j,s}$, with $1\le i\le 17, 1\le j\le 40, 1\le s\le 2$, on the modularized training images $B_{i,j,s}$, This is done by first tranfsforming the original $64\times 64$ color images into the input vectors of the first layer of the FNN, and then obtaining the predicted labels of the FNN. 

The transformation first down-samples the images to $32\times 32$, using the mean values of non-overlapping $2\times 2$ grid cells. Then it trims off one row from the top and bottom, and,  respectively, one column from the left and right borders of the images. Featured images are obtained from the primary color channel images according to the combination weights from Tab.\ref{tab_feature}. Each featured image is then scaled to cell values between $[-1,1]$. As a result, the final $30\times 30$ image matrix is changed to a 900-dimensional input vector for the FNN. The image transformation is performed by the function code, fun\_transform\_data\_rgbfeatures.m. In the Data\_Transformation folder, the code script, Run\_Transform\_Data.m, transforms the modularized dataset to the vectorized input for all training FNNs.  

\begin{table}[t]
    \centering
    \begin{tabular}{c|ccccc}
     proto-model   &   min   &   mean   &   median  &   max     &  100rt stat.$\quad$ \\
     \hline
    \text{T\_h1\_m1}  &  98.965 &   99.923 &   99.975  &     100   &   196/1360,\ 12/80  \\    
    \text{S$'$\_h1\_m1}  &  98.866 &   99.815 &   99.876  &  99.981   &   --    \\
    \text{T\_h1\_m2}  &  98.736 &   99.919 &   99.969  &     100   &   196/1360,\ 12/80   \\   
    \text{S$'$\_h1\_m2}  &  98.723 &   99.831 &   99.888  &  99.994   &  --      \\
    \text{T\_h2\_m1}  &  98.884 &   99.898 &   99.956  &     100   &   154/1360,\ 10/80   \\  
    \text{S\_h2\_m1}  &  98.745 &   99.783 &   99.856  &  99.987   &   --      \\
    \end{tabular}
    \vskip .2in
    \caption{Training accuracy rates. All statistics are calculated against the total number of $N_{i,j,s}$, 1360. The last column is for the error-free FNNs, distributed over the total number of modules and their subsets, 80. For the T\_h1\_m1, for example, the number of error-free FNNs per module is 196/12 = 16.3333. That is, there are 11 modules of which all 17 features are trained error-free, and one of the 12 modules has 9 out 11 features that are trained error-free. The missing two feature FNNs can be made error-free by the GDT algorithm with stricter search settings.}
    \label{tab_training_stat}
\end{table}

Specifically, all SGD proto-models are trained with a threshold stop at 98\%. Above the threshold, the GDT training follows right afterward to obtain the corresponding GDT proto-models. The SGD proto-model with two hidden layers, S\_h2\_m1, presented is of this throttling-down type. Figure \ref{fig_wheel} gives an illustration for these two types of proto-model conjoining training. It is chosen to show the dependency relationship between SGD and GDT trainings when the data batch $B_{i,j,s}$ is free of double labeling. In contrast, for the SGD proto-models with one hidden layer, S\_h1\_m$y$, $y=1,2$, after the GDT proto-models are trained, the threshold break for the SGD training is released, and we then train the SGD proto-models in a few more gradient descent iterations to a higher accuracy, allowing the training algorithm to realize its full potential. We use S$'$\_h$x$\_m$y$ to denote this SGD training type, and $\{W'_s,b'_s\}$ for their trained parameters. 

Table \ref{tab_training_stat} lists the performances for all types of training, in terms of the minimum, mean, median, and maximum accuracy rates, and the number of featured modules that reach  100\% accuracy. For example, the result shows that of the 80 module partitions, 12 are free of doubling labeling. The minimum accuracy for the GDT training gives a lower estimate for the scope of the double-labeling problem. It also shows that the GDT proto-models always perform better than the SGD proto-models, with or without the SGD threshold break. The code, Run\_Training\_Analysis.m, is used to produce this table. 

\begin{table}[t]
    \centering
    \begin{tabular}{c|cc|cc|cc}
featured & \multicolumn{2}{c|}{\text{T\_h1\_m1}} & \multicolumn{2}{c|}{\text{T\_h1\_m2}} & \multicolumn{2}{c}{\text{T\_h2\_m1}} \\ 
model & Acc. & \text{Top-1} & Acc. & \text{Top-1} & Acc. & \text{Top-1} \\ 
 \hline
\text{Model-1} &      86.848  &    99.998  &    86.047  &    99.999  &    73.065  &   100 \\
\text{Model-2} &      95.931  &    99.986  &    95.767  &    99.986  &    92.900   &   99.954 \\
\text{Model-3} &      97.382  &    99.982  &    97.275  &    99.980  &    95.979  &    99.866 \\
\text{Model-4} &      97.897  &    99.978  &    97.879  &    99.978  &    97.014  &    99.802 \\
\text{Model-5} &      98.133  &    99.979  &    98.126  &    99.970  &    97.487  &    99.755 \\
\text{Model-6} &      98.289  &    99.977  &    98.300  &    99.969  &    97.770  &    99.729 \\
\hline
           mean &      95.747  &    99.983  &    95.566  &    99.981  &    92.369  &    99.851 \\
\hline
 & \multicolumn{2}{c|}{\text{S$'$\_h1\_m1}} & \multicolumn{2}{c|}{\text{S$'$\_h1\_m2}} & \multicolumn{2}{c}{\text{S\_h2\_m1}} \\ 
\text{Model-1} &      88.728 &   99.987 &   89.191 &   99.991 &   72.925 &   99.992 \\
\text{Model-2} &      96.282 &   99.757 &   96.371 &   99.796 &   91.570 &   99.715 \\
\text{Model-3} &      97.444 &   99.509 &   97.486 &   99.596 &   94.881 &   99.318 \\
\text{Model-4} &      97.905 &   99.309 &   97.953 &   99.429 &   96.068 &   98.991 \\
\text{Model-5} &      98.106 &   99.143 &   98.168 &   99.275 &   96.633 &   98.734 \\
\text{Model-6} &      98.247 &   99.008 &   98.299 &   99.166 &   96.990 &   98.569 \\
\hline
           mean &      96.119 &   99.452 &   96.245 &   99.542 &   91.511 &   99.220 \\
    \end{tabular}
    \vskip .2in
    \caption{Accuracy rates for featured models. Notice that Model-6 is identical to its proto-model because both contain all the training features.}
    \label{tab:model_stat}
\end{table}

\medskip\noindent
\textbf{Model Evaluation.} As discussed above, when an image is fed to the model, the model must run it through all featured FNNs against the label classes 0 through 999 before making its prediction by the majority-vote protocol. For a single or a small batch of images, this is done by the code, Ru\_Evaluation\_Model\_Individual.m, in the Model\_Evaluation folder. But, for the entire dataset of 1.2 million plus images, this program is too slow. Instead, a parallelized code, Run\_Evaluation\_main\_Model.m, is used to run on batches of the dataset to speed up the process. It outputs the predicted labels of the dataset for the feature-defined Model-1 through Model-6, introduced in the previous section. After the predictions are saved, we can run two more files. One file, Run\_Evaluation\_post\_Model.m, finds the accuracy rates of all six models, and the Top-1 accuracy rate, which is defined to be the rate by which a correct prediction is made by the super majority of the votes, namely the first choice of the predictions. The other file, Run\_Evaluation\_Confusion\_Matrix.m, builds the confusion matrices for the models, from which we can find the labels whose entire image classes are classified perfectly. All these tasks for all featured models, derived from the two types of proto-model can be accomplished by the code, Run\_Model\_Analysis.m in the ImageNet-1k folder.  The results are shown in Tab.\ref{tab:model_stat}. It shows the following: The more features a model has, the more accurate it becomes, with a modest drop in the Top-1 rate. Models with a single hidden layer perform better than models with two hidden layers. The Top-1 rate for the GDT-trained featured models is always better than the SGD-trained featured models. For the conjointly trained models for which GDT training is directly based on SGD-trained models, featured models of the former always perform better than models of the latter type, as shown by the h2-models. However, when this dependence is removed, i.e., the SGD proto-models are trained beyond the 98\% threshold stoppage, their offspring featured models can outperform the GDT kind in model accuracy, but not in the Top-1 accuracy.  

\begin{table}[t]
    \centering
    \begin{tabular}{c|cc|cc|cc}
     & \multicolumn{6}{c}{Errorless Trained Labels}\\
featured & \multicolumn{2}{c|}{\text{T\_h1\_m1}} & \multicolumn{2}{c|}{\text{T\_h1\_m2}} & \multicolumn{2}{c}{\text{T\_h2\_m1}} \\ 
model & by batch & all & by batch & all & by batch & all \\ 
 \hline
\text{Model-1} &          0.8 &       0 &     1.4 &       0 &       0 &       0 \\
\text{Model-2} &         59.7 &       0 &    46.9 &       0 &     4.6 &       0 \\
\text{Model-3} &        153.7 &       1 &   131.2 &       1 &    38.5 &       0 \\
\text{Model-4} &        220.9 &       7 &   201.8 &       5 &      90 &       0 \\
\text{Model-5} &        256.8 &       9 &   245.6 &       5 &   128.9 &       0 \\
\text{Model-6} &        285.9 &      11 &   278.9 &       8 &   168.3 &       1 \\
    \end{tabular}
    \vskip .2in
    \caption{Confusion matrix statistics. Here is what the numbers mean. For example, for Model-6 of T\_h1\_m1, the `by batch' number is the average number of labels that are perfectly classified over the 10 batches of the ImageNet-1k dataset, which is 285.9. In comparison, when aggregated over the 10 batches, the number drops to 11 labels, in the `all' column. The corresponding labels are label-number 6, 293, 386, 420, 451, 467, 567, 640, 642, 644, and 708.}
    \label{tab:conf_stat}
\end{table}

Table \ref{tab:conf_stat} summarizes some statistics of the confusion matrices. The code, Run\_Confusion\_Matrix\_Analysis.m, generates all the numbers in the table. The main finding is that the more features a model has, the more perfectly classified labels the model is capable of gaining. It also shows that having one additional hidden layer degrades the outcome in all the evaluation categories analyzed in this paper. The same script can be used to find the statistics for all featured models of the SGD and GDT types.

\begin{table}[t]
\centering
\begin{tabular}{c|cc|cc|cc}
featured & \multicolumn{2}{c|}{\text{T\_h1\_m1}} & \multicolumn{2}{c|}{\text{T\_h1\_m2}} & \multicolumn{2}{c}{\text{T\_h2\_m1}} \\ 
model & Acc. & \text{Top-1} & Acc. & \text{Top-1} & Acc. & \text{Top-1} \\ 
 \hline
\text{Model-1} &      74.843  &       100  &    72.427  &       100  &    41.807  &       100 \\
\text{Model-2} &      94.755  &       100  &    94.456  &       100  &    84.018  &       100 \\
\text{Model-3} &      97.012  &       100  &    96.921  &       100  &    92.438  &    99.997 \\
\text{Model-4} &      97.714  &       100  &    97.728  &       100  &    94.807  &    99.992 \\
\text{Model-5} &      98.026  &       100  &    98.022  &    99.998  &    95.939  &    99.987 \\
\text{Model-6} &      98.213  &    99.999  &    98.224  &    99.999  &    96.537  &    99.983 \\
\hline
           mean &    93.427  &   99.999  &   92.963  &   99.999  &   84.258  &   99.993 \\
\hline
 & \multicolumn{2}{c|}{\text{S$'$\_h1\_m1}} & \multicolumn{2}{c|}{\text{S$'$\_h1\_m2}} & \multicolumn{2}{c}{\text{S\_h2\_m1}} \\ 
\text{Model-1} &      80.573  &       100  &    80.435  &       100  &    45.154  &       100 \\
\text{Model-2} &      95.608  &    99.995  &    95.743  &    99.997  &    81.292  &    99.991 \\
\text{Model-3} &      96.952  &    99.949  &    97.093  &     99.960  &    89.632  &    99.925 \\
\text{Model-4} &      97.383  &    99.851  &     97.570  &    99.876  &    92.371  &    99.804 \\
\text{Model-5} &      97.549  &    99.741  &    97.748  &    99.757  &    93.758  &    99.674 \\
\text{Model-6} &      97.669  &    99.635  &    97.866  &    99.663  &    94.569  &    99.562 \\
\hline
          mean &      94.289  &   99.862  &   94.409  &   99.876  &   82.796  &   99.826 \\
\end{tabular}
\vskip .2in
\caption{Accuracy rates for the expanded majority voting protocol.}
\label{tab:model_stat_2}
\end{table}

Specifically, the majority voting protocol for each featured model consists of three voting steps. Let $p$ be the number of features for the model. For Model-1, $p=3$; for Model-2, $p=6$, and so on. First, for each input image, we choose the top label output from every FNN $N_{i,j,s}$ of the featured model, where $i$ ranges from 1 to the model defining $p$ features. In the second step, the selection is carried out by the module-submodule, $n_{j,s}$, over the set of $p$ features, resulting in $kr=80$ selected labels together with the number of votes each winning label gets. In the ideal situation where all FNNs are perfectly trained, the module-submodule that is assigned to train the image will return the true label together with the super-majority vote number $p$. In contrast, for any other module-submodule, its selected label's winning number is between 1 and $p$. It is possible that such a wrong label can have the super-majority winning number $p$. In the third step, selection is carried out for the set of winning numbers. Again, for the ideal situation, if no other module-submodules reach the super-majority winning number $p$, the final output will be the true label by the module-submodule that is trained on the image. Otherwise, if there is a tie because a wrong label also has the super-majority number $p$ by chance, we first compute the variance of the loss function values over the features for each of the tying module-submodules, and then choose the module-submodule having the smallest variance to break the tie. Table \ref{tab:model_stat} is produced by this procedure. 

Table \ref{tab:model_stat_2} is produced the same way except for the following modification for the first step. Rather than choosing only the top label output from every FNN $N_{i,j,s}$ of the featured model, we use instead the top three labels from every FNN, with the effect of increasing the pool of candidate labels for the first round of selection. This is because the intended label may not always be the top choice of a given feature, but is more likely among the top two or top three choices. That is, the bigger the pool, the more likely the true label will reach the super-majority. The results show that the Top-1 rate improves considerably with a minor drop in the overall accuracy. For example, for Model-5 of T\_h1\_m1, every correctly predicted label is by the intended super-majority, namely, the winning number of 15 votes. If the dataset is consistent without double labeling, the accuracy rate should be significantly higher, possibly to 100\%, the same as the Top-1 rate. 

\section{Discussion}
The continuous UAT (\cite{Hornik1989, cybenko1989approximation}) holds a deep theoretical interest on its own. But, the discrete UAT is extremely important for applications because all practical problems can be treated as either discrete directly or approximated as discrete with the approximating error smaller than the UAT's discriminating threshold. That is, all functions can be represented exactly by FNNs by the discrete UAT. The GDT method is the only known method capable of realizing this potential. Models that can be trained error-free are models that can truly improve on themselves. 

The analysis and results from previous sections suggest that for datasets free of double labeling, the architecture with $k=r=1$ always gives rise to perfectly trained models. This can be the desired outcome if the dataset is not too big, such as the MNIST, and MedMNISTs from \cite{yang2023medmnist}. But for large datasets, such as the ImageNet-1k considered here, training a single FNN without modularization becomes prohibitive in terms of computer memory size and running time because the number of training images and the size of the output layer are both substantial. This is the primary reason why we adopted the modularized architecture for the model. In addition to the intended advantage of only requiring modest computing capabilities for training of modularized FNNs, model updating can be carried out by retraining the modules that need updating. For large datasets, we trade some loss of accuracy for efficiency by using modularization. 

For our model, the modules are organized in the numerical order of the dataset's labels. However, a more rational approach should be taken, similar to the idea of taxonomy. For example, we can group animals, plants, etc., in different modules. Another idea is to construct expert models, each for a distinct category of objects. The obvious advantage is efficiency because we do not need to check a known type of image against irrelevant types. Another obvious advantage is cost savings for retraining, for which only the relevant modules are updated. Datasets similar to ImageNet-1k, in our view, should be organized by categories.  

For model architecture, our results suggest that having more hidden layers does not improve model performance. As a probable explanation, the nonlinear transformations of the additional layers may have erased some information essential for errorless classification. This finding seems consistent with the discrete version of UAT, which requires only one hidden layer for errorless classification. However, for datasets free of double labeling, GDT guarantees 100\% accuracy regardless of the number of hidden layers.  

We did not reach our goal of training the ImageNet-1k to perfection. The problem may have an explanation because we limited all the model parameters to 4-decimal-place accuracy. Although it is a possibility, it is unlikely because when our GDT algorithm runs at its maximum strength, the error-free trained modules are very uniform: either all features of a module are trained error-free, or no feature of a module can reach 100\% accuracy, suggesting that some modules are free of double labeling and some others are not. In addition, the trained models with the given decimal place precision are robust with respect to the model parameters because all the results remain the same quantitatively when the parameters are truncated to the 3rd decimal place. As a result, we strongly believe that the ImageNet-1k may contain double-labeled images. However, it is not our goal to make a definitive determination. 

Assuming that is the case, here is an estimate on the scope of the problem. From the minimum accuracy rates of the GDT proto-models in Tab.\ref{tab_training_stat}, we can obtain an estimate for the maximum error rate around 0.0115, which translates to an estimated 14,000 images. Assuming only 10 percent of these images that is ultimately responsible for the double labeling problem, we are left with a set of 1,400 images. Averaged over 1000 labels, it suggests that there are 1 to 2 images per label on average that are double labeled, looking like the result of human curating error. An algorithm can be developed based on our GDT search algorithm to find the errant labels. This can be the subject of a future study. 

If double labeling is detected in a dataset, we should be extra cautious when deploying a model trained on it. This is because the problem is bound to cause classification errors, which can cascade down to applications. Specifically, for datasets free of contractions, the accuracy rate for GDT-trained models in Tab.\ref{tab_training_stat} should be 100\%. Otherwise, the accuracy statistics for featured models in Tab.\ref{tab:model_stat} will be muddled up. Due to these reasons, we did not pursue a model evaluation on the validation data since our search algorithm is not equipped to isolate the causes of the prediction error. The dataset's probable curating errors can cause needless inaccuracy for the GDT-trained models, which will ripple through the featured models. As a result, we used only the simpler kinds of features, consisting of linear and reversible combinations of the primary color channels, to test the GDT method on the dataset. 

For image datasets free of double labeling, our GDT method can train models error-free, a vital goal in some applications, especially in the fields of Medicine and Healthcare, where repeating errors must not be tolerated. For problematic image datasets, our results suggest that it may not be worthwhile to add GDT training to SGD training, because the overall performance of our SGD models is on par with the GDT models, and GDT training is considerably more time-consuming than SGD training. We note that our SGD models with six features (Model-2),  113,798,880 parameters, and 566,880 neurons, have already outperformed all conventional SGD methods in the literature.  
For practical applications and image datasets free of double labeling, we envision the inclusion of more sophisticated features, such as various convolutions of, or singular value decompositions of images. Such feature inclusions are for future studies to quantify the predictive capability of trained models in terms of the size of neighborhoods of training data in which models' predictions remain true. Such studies should have a more concise explanation for models' performance on validation and test data, as well as the true scope of models' capabilities and limitations. In addition, only for consistent datasets can we pursue the smallest FNN model that can be trained fully. For the MINST handwritten digits, we only need one hidden layer of 20 neurons and 15,910 parameters to achieve that goal (\cite{deng2023error-free}). We believe that the size of our model for ImageNet-1k can be reduced by one order of magnitude for the same level of performance, which we did not pursue in this study.    

The tree-like structure of our model is similar to taxonomy. We believe that datasets, such as species taxonomy, the catalog of the Library of Congress, or any database requiring precision in bookkeeping, can be represented by AI models like ours. Most importantly, if such a model makes a mistake, the error-free training method can ensure that an updated module will not make the same mistake. Errorless training can be beneficial for human endeavors that require predictive precision AI.  

Because our model is structured in parallel FNNs with minimal hidden layers and modest batch sizes, training is limited only by the number of CPU cores. In the ideal setting where each FNN is assigned to a core, an SGD proto-model can be trained in 1 hour, and a GDT proto-model can be trained in 3 hours, which is the same length of time to update any FNN. As a demonstration, no GPUs were used to train our model. However, we expect that our current training speed can be considerably boosted by incorporating GPU parallelism into our training code.  

\medskip\noindent
\textbf{Acknowledgment:} All large-scale computations for the paper were carried out in the Holland Computing Center of the University of Nebraska System. Thanks to Lucas Wurtz for his enthusiasm to get the project started. 

\medskip\noindent
\textbf{Data Availability:} All model weights and evaluation codes can be found in the Hugging Face repository, \cite{Repo_Towards_Errorless_Training_ImageNet-1k}. For questions about the training methods, send emails to the first author. 
 
\bibliographystyle{abbrv}
\bibliography{bibliography}


\end{document}